\documentclass[conference]{IEEEtran}
\IEEEoverridecommandlockouts
\usepackage{graphicx}
\usepackage{latexsym}
\usepackage{amsmath,amssymb,amsfonts}
\usepackage{booktabs}
\usepackage{url}
\usepackage{adjustbox}
\usepackage{todonotes}
\usepackage{hyperref}
\usepackage[noadjust]{cite}
   \hypersetup{
     colorlinks=true,
     linkcolor=blue,
     filecolor=blue,
     citecolor = black,      
     urlcolor=blue,
     }

\title{Unsupervised Learning of Explainable Parse Trees for Improved Generalisation}
 
\author{ Atul Sahay \hspace{0.3em} Ayush Maheshwari \hspace{0.3em} Ritesh Kumar \hspace{0.3em}\\  Ganesh Ramakrishnan \hspace{0.3em} Manjesh Kumar Hanawal \hspace{0.3em} Kavi Arya \\
Indian Institute of Technology Bombay\\
{\tt \{atulsahay,ayusham,riteshkumar,ganesh,kavi\}@cse.iitb.ac.in, mhanawal@iitb.ac.in}
}
\begin{document}
\maketitle
\begin{abstract}
 Recursive neural networks (RvNN) have been shown useful for learning sentence representations and helped achieve competitive performance on several natural language inference tasks. However, recent RvNN-based models fail to learn simple grammar and meaningful semantics in their intermediate tree representation. 
In this work, we propose an attention mechanism over Tree-LSTMs to learn more meaningful and explainable parse tree structures. We also demonstrate the superior performance of our proposed model on natural language inference, semantic relatedness, and sentiment analysis tasks and compare them with other state-of-the-art RvNN based methods. Further, we present a detailed qualitative and quantitative analysis of the learned parse trees and show that the discovered linguistic structures are more explainable, semantically meaningful, and grammatically correct than recent approaches. The source code  of the paper is available \href{https://github.com/atul04/Explainable-Latent-Structures-Using-Attention}{here}. \footnote{Accepted at IJCNN 2021. To be governed by IEEE Copyright.}
\end{abstract}

\section{Introduction}
Distributed word representations are well known and widely used in natural language processing for solving downstream tasks ~\cite{pennington-etal-2014-glove} such as sentiment analysis, summarisation, semantic matching, and more. Additionally, sentence or phrase representation models such as Long-Short Term Memory (LSTM)~\cite{10.1162/neco.1997.9.8.1735} use sequential data combining current word state with previous states. However, these recurrent models do not necessarily comply with the grammatical structure of the text, thereby adversely affecting the sentence representation.
Based on linguistic theories that have promoted the use of constituent tree-based representation of natural language text, tree-based models such as RvNN (Recursive Neural Networks) have been proposed to learn sentence representations from syntactic parse trees \cite{socher-etal-2011-semi}. These models rely on structured input, i.e. parse trees to encode sentences recursively from leaf nodes to the root node of the tree. RvNNs are generalised recursive neural networks (RNN) that operate over left- or right-skewed trees, while RNNs work with linear chain structures.
\par
Due to the RvNN's ability to encode sentences and capture semantics from parse trees, they work effectively on tasks such as Natural Language Inference (NLI) or text entailment. The NLI problem is to determine whether a hypothesis sentence can be inferred from the premise sentence.

\begin{figure}[ht!]
    \centering
    \includegraphics[height=4.5cm, width=0.48\textwidth]{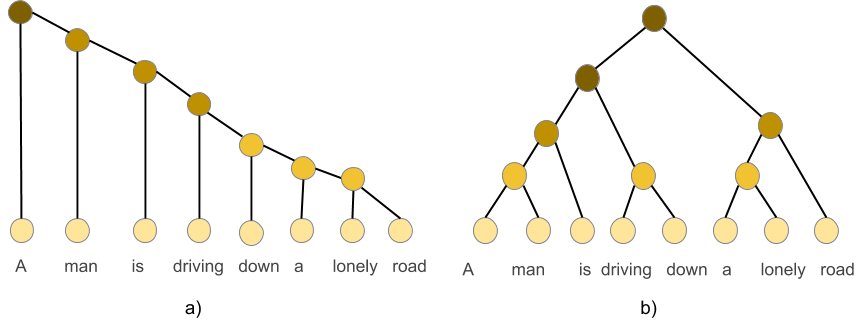}
    \caption{For a given example ~\texttt{A man is driving down a lonely road} a) shows an output of parse tree from Gumbel Tree-LSTM  and b) output using our approach.}
    \label{fig:example}
\end{figure}
 Consider the sentences: A:\texttt{ An older and younger man smiling.}, B:\texttt{ Two men are smiling and laughing at the cats playing on the floor.} and C:\texttt{ Some men are laughing.} Statement A is a premise and, B and C are the hypotheses. If statement B or C can be inferred from A, then the output label is \textit{entailment}. If B or C contradicts A, then output label is a \textit{contradiction}, and if nothing can be determined, then \textit{neutral}. In the above example, statements A and B are neutral while A and C have an entailment relationship.
NLI task relies on efficient learning of parse trees for determining entailment relationship. However, the parse tree has a high annotation cost as it requires a significant amount of expert-level supervision. 
\par

Several approaches  do not need annotated data to learn parse trees, such as Gumbel Tree-LSTM \cite{gumbel}. It uses a composition vector to recursively select nodes until a single node vector remains at the end. However, Gumbel Tree-LSTM gives uniform weightage to each node of the tree and composes task-specific tree structure. In this work, we propose an attention mechanism that leverages latent information of the parse tree in an unsupervised manner.
Figure \ref{fig:example} shows the Gumbel Tree-LSTM do not adequately capture the intermediate structure (parse trees) and produces skewed trees. On the other hand, our method produces improved intermediate parse tree structures.
\par

Inspired by the Tree LSTM based models, we adopt an attention mechanism that generates better intermediate structures that encourage the model to focus on salient latent information of the parse tree relevant for the classification decision.
Another motivation is the principle of compositionality as studied by Frege~\cite{pelletier1994principle} (as well as by ancient grammarians such as P\=a\d nini). The principle of compositionality  states that the meaning of any complex expression is determined by the meaning of its constituent parts and syntactic rules used to combine them. \par
From experiments on natural language inference, semantic relatedness, and sentiment analysis task, we find that our model outperforms state-of-the-art RvNN-based models. Further, we evaluate the induced tree quantitatively and qualitatively and observe that induced structures follow grammatical principles and generate more meaningful structures than previous approaches. 

\par
Our contributions can be summarised as follows:
\begin{itemize}
    \item We propose a novel attention mechanism over Tree-LSTM that assigns differential weightage to each item in the parse tree.
    \item We conduct extensive experiments on natural language inference, semantic relatedness, and sentiment analysis tasks and outperform state-of-the-art RvNN-based methods.
    \item We quantitatively and qualitatively demonstrate that our approach learns more explainable, semantically meaningful, and grammatically correct parse tree structures.
\end{itemize}
 \section{Related Work}
Tree-structured recursive neural networks (RvNN)~\cite{socher-etal-2011-semi} builds a vector representation for a sentence by incrementally computing representations for each node in its parse tree have been proven to be effective at sentence understanding tasks such as  sentiment analysis~\cite{socher-etal-2013-recursive}, textual  entailment~\cite{Bowman_2016}, and translation ~\cite{eriguchi-etal-2016-tree}. 

However, these methods are domain-dependent, slow, and error-prone and incur high annotation costs since they are supervised approaches requiring parse-tree annotations. Recently, several unsupervised approaches \cite{yogatama2016learning,choi2017learning,Maillard_2019} have been proposed that learn supervision from NLI tasks to learn parse trees. However, these methods cannot learn simple grammar and meaningful semantics, though they perform well on NLI tasks \cite{williams2018latent}. Additionally, several approaches \cite{drozdov2019unsupervised, kim2020compound, sdiora} aim to learn unsupervised parse trees; however, they perform poorly on end task. In this paper, we demonstrate that our approach can capture both grammar and semantics in the sentence, thus learning better parse trees and outperform RvNN-based model on several tasks.
\par
RL-SPINN \cite{yogatama2016learning} used reinforcement learning paradigm to learn tree structures and employ their performance on the downstream task as a reward to train the algorithm. They leverage the SPINN architecture \cite{Bowman_2016} and apply TreeLSTM as a \textit{reduce} composition function. Owing to their reinforcement learning-based approach, their model takes time to train and is not easily extensible to multiple tasks. Moreover, as we will show in our experimental section, the grammar or meta-level association thus detected by their approach is relatively trivial. Recently, \cite{drozdov2019unsupervised} proposed an unsupervised latent chart tree parsing algorithm, {\em viz.}, DIORA, that uses the inside-outside algorithm for parsing and has an autoencoder-based neural network trained to reconstruct the input sentence. DIORA is trained end to end using masked language model via word prediction.  The chart filling procedure of DIORA is used to extract binary unlabeled parse trees. It uses the CYK algorithm to find the maximal scoring tree in a greedy manner. As of date, DIORA is the state-of-the-art approach to unsupervised sentence parsing. Though their approach produces better parse trees; however it performs extremely poorly on NLI tasks.
\cite{Maillard_2019} presented CYK parsing algorithm that computes $O(N^2)$ tree nodes for $N$ words in the sentence. Though this model is easily trainable using a backpropagation algorithm, it is memory intensive since possible tree nodes increase linearly with depth. Additionally, the structures predicted by them are ambiguous and harder to interpret. In order to prevent parsing all possible trees, \cite{choi2017learning} proposed straight through Gumbel softmax \cite{gumblesoftmax} to estimate gradients vector without parsing all trees. However, their model produces non-meaningful syntactic and semantics of the sentence \cite{williams2018latent}. 
\par

\section{Our Model}
Given an input sentence, our model learns the parse tree by applying composition functions using Gumbel Tree-LSTM.

\begin{figure*}
    \centering
    \includegraphics[width=\textwidth]{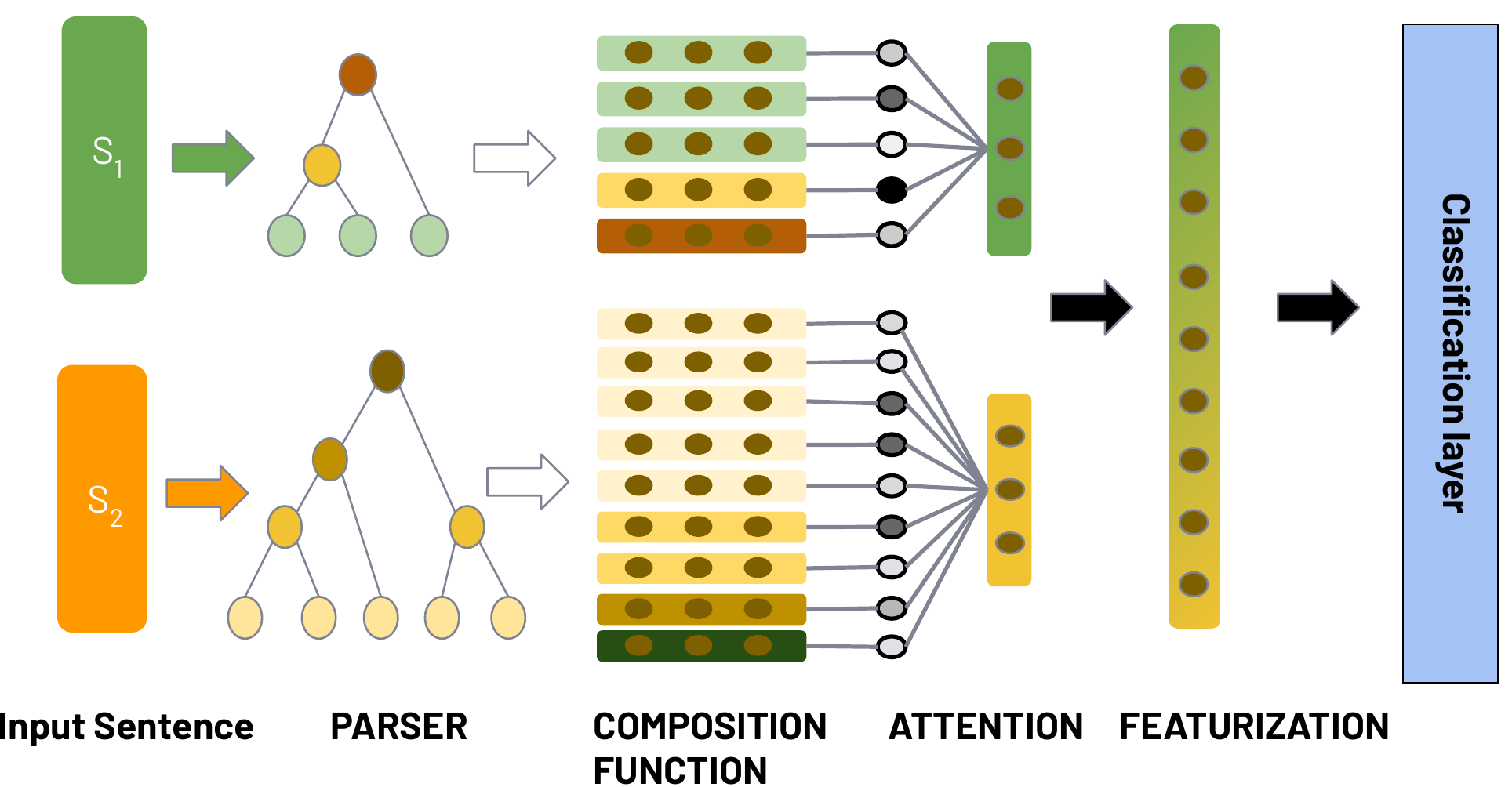}
    \caption{Architecture of our approach on sentence pair classification tasks, the $S_1$ and $S_2$ represents input word embeddings for sentences, \textbf{PARSER}: The bottom up parser induces a parsing for the sentence giving node representation, the \textbf{COMPOSITION FUNCTION} uses each node representation to obtain the sentence representation using \textbf{ATTENTION}, the \textbf{FEATURIZATION} further transforms the obtained embeddings from previous layers which is fed to the \textbf{classification layer}.  }
    \label{fig:arch}
\end{figure*}

We learn to attend over the output of the composition function for classification that enables our model to dynamically compose an unlabelled parse tree in a bottom-up manner. Finally, sentence representations are transformed to form a feature vector used for the downstream classification task. In subsequent subsections, we present the details of each component. Our proposed architecture is shown in Figure \ref{fig:arch}.

\subsection{Gumbel Tree-LSTM}
Given an input sequence and its associated directed acyclic graph (DAG), Tree-LSTM applies transformation to the representations of the lowest level leaf nodes of the tree. The composition function merges these transformation for next level of nodes in the DAG. Tree-LSTM generalises the LSTM cell to tree-structure and formulates its composition function as 
\begin{gather}
    \begin{bmatrix} z \\ i \\ f_l \\ f_r \\o \end{bmatrix} = 
    \begin{bmatrix}
    tanh \\ \sigma \\ \sigma \\ \sigma \\ \sigma
    \end{bmatrix}  \Big( \mathbb{R} \begin{bmatrix}
    h_l \\ h_r 
    \end{bmatrix} + b \Big)
\end{gather} 
\begin{align}
    & c_p = z \odot i + c_l \odot f_l + c_r \odot f_r \\
   & h_p = tanh(c_p) \odot o
\label{eq:treelstm}
\end{align}
where $f_l$ and $f_r$ are the forget gate for left and right nodes of the tree, an output gate $o$, a memory cell $c$, a hidden state $h$, 
$\mathbb{R}^{5D \times 2H}$ is the recursion function, $\sigma$ denotes the sigmoid function, $tanh$ is hyperbolic tangent function and $\odot$ is the element-wise product. 
\par

Gumbel Tree-LSTM, based on Tree-LSTM, generates the parse tree in an unsupervised manner. It is motivated by Gumbel max trick \cite{gumblemaxtrick}, an algorithm for sampling categorical distribution using  the discontinuous \texttt{argmax} function. It uses Straight-through (ST) Gumbel softmax estimator \cite{gumblesoftmax} to sample compositions during training. Gumbel softmax replaces \texttt{argmax} with the differentiable \texttt{softmax} function. ST Gumbel softmax is a discretized version of Gumbel softmax, and similar to straight-through estimator \cite{ste}. 
\begin{align}\nonumber
  u_i \sim Uniform(0,1)  
\end{align}
\begin{align}
\epsilon_i = -\log(\log(u_i) )
\label{eq:gumble}
\end{align}
Gumbel noise, $\epsilon_i$,  is a perturbation such that \texttt{argmax} is equivalent to drawing a sample weighted average. In the forward pass, it discretizes the continuous output vector $v$ from Gumbel softmax distribution into the one-hot vector $z$. 
\begin{align}
z = \arg \max(v + \epsilon)   
\label{forward}
\end{align}
and in the backward pass, it uses continuous $p$, obtained from the continuous output vector $v$.
\begin{align}
p = softmax(v + \epsilon)
\label{backward}
\end{align}

\par
Initially, Gumbel Tree-LSTM transforms an input leaf node vector $i$ to a pair of vectors $r_i^0 = (h_i^0, c_i^0)$. The initial state $r_i^0$ can be transformed using methods such as affine transformation, LSTM and bi-directional LSTM transformation. The tree is built in bottom-up manner by recursively merging consecutive constituents using Eq. (\ref{eq:treelstm}). The constituents are merged on the basis of validity score of the representation defined by $q.h$, where $q$ is a compositional trainable query vector. At each layer $t$, the model calculates the normalised validity score $v$ of each candidate $M_i^{t+1}$, that consists of composing two consecutive words on the $t+1^{th}$ layer using
\begin{equation}\nonumber
    v_i = \frac{\exp (q.h_i^{t+1})}{\sum_{j=1}^{t+1} \exp(q.h_j^{t+1})}
\end{equation}
The merge procedure chooses to compose a candidate with the highest validity score revalued after adding Gumbel noise, $\epsilon_i$. The procedure is then repeated until a single representation remains at the end. Its hidden state is the constituent tree representation. The advantage of Gumbel softmax is that training objective becomes differentiable. However, it introduces bias in the gradient estimates due to random restarts. The approximation based on Gumbel-softmax distribution makes it tricky to recover simple context-free grammars \cite{nangia2018listops}.

\subsection{Structural Attentive Layer}
In Gumbel Tree-LSTM, a uniform composition weightage is given to each node of the tree or the constituents. However, constituents should be weighted and merged according to its latent part of speech information. By attentively weighting and combining all intermediate tree representations, the model provides multiple paths for gradients to flow back to the input, thus improving input representations. Additionally, the attention allows the model to focus gradient updates to more important merging decisions in the lower levels of the tree in the backward pass.
We utilise an attention mechanism that aims to encourage the model to focus on salient latent information of the composition that is appropriate for the end task.
We denote the output of the intermediate nodes and leaf nodes in the layers of the Tree-LSTM as $(\overrightarrow{h_1}, \overrightarrow{h_2}, \cdots , \overrightarrow{h_{2n-1}}) $ where $n$ is the sentence length.
The formulation of the architecture is as follows :
        \begin{equation}
        {e_i}^l = ReLU\left ( W_e\times\vec{{h_i}} \right )
        \label{embdacq}
        \end{equation}
        \vspace{-0.5cm}
        \begin{equation}
        \tilde{{a_i}} = \exp(W_a\times{e_i})
        \end{equation}
        \begin{equation}
        {a_i} = \frac{\tilde{{a_i}}}{\sum_{j=1}^{2n-1}(\tilde{{a_j}})}
        \end{equation}
        \begin{equation}
        \vec{H_c} = \sum_{i=1}^{2n-1} ({a_i}\dot{\vec{{h_i}}})
        \label{context}
        \end{equation}

 We leverage the relationships among the different nodes of the tree that can learn a shared embedding space. Each such hidden state representation of tree nodes viz., $(\overrightarrow{h_1}, \overrightarrow{h_2}, \cdots , \overrightarrow{h_{2n-1}})$ maps to lower dimensional space $\mathbb{R}^{D_a}$. This step in Eq.~\ref{embdacq} ensures that every node vector is brought to the shared embedding space or embedding acquisition \cite{yogatama}. A more sensible and finer exploration of meta-level associations can be made by capturing relationships between different nodes in the shared embedding space through our attention mechanisms. The transformed nodes vectors are denoted by $e_i \in \mathbb{R}^{D_a \times 1}$ with weight matrix  $ W_e  \in  R^{D_a \times H}$ associated to embedding acquisition layer. A scalar unit $a_i$ is learnt with the weight matrix $ W_a  \in  R^{1 \times D_a}$. These scalar units are then normalised to 1, and the whole formulation of context word vector is defined in Eq. \ref{context}. The context vector $H_c$ is learned using attention weighted contextual representation by passing node vectors to the fully connected layer with $D_a$ units in a hidden layer. 
\par 
In the last step, we concatenate learned sentence representation, $s_1$ and $s_2$ obtained from forward pass of sentence 1 and sentence 2 respectively with the $|s_1-s_2|$ and $s_1 \odot s_2$. The final feature vector (for sentence pair classification tasks), $f \in R^{4 \times D_a}$ is: 
$$f = [s_1,~ s_2, ~|s_1-s_2|, ~s_1 \odot s_2] $$  where $\odot$ is the element-wise product. This feature vector forms as input to the final classification layer.
\section{Experiments}
We perform experiments on three benchmarks tasks: NLI task on Stanford NLI (SNLI) dataset, semantic relatedness task on Quora Question Pairs (QQP) dataset, and sentiment classification task on Stanford sentiment treebank (SST) dataset. 

\begin{table}[h]
 \caption{Results on SNLI dataset. $^{\dagger}$: results are taken from \cite{williams2018latent}, $^{\mathsection}$: results reported in \cite{ppo} and $^{\ast}$: publicly available code and hyperparameter optimization was used to obtain results. PPO: Proximal Policy Optimization}
 \centering
 \begin{adjustbox}{width=0.9\linewidth}
\begin{tabular}{@{}lcc@{}}
\toprule
Model & Dim. & Acc.\\
\toprule
RL-SPINN \cite{yogatama2016learning}   & 100D & 80.5 \\
Unsupervised Tree-LSTM \cite{Maillard}   & 100D & 81.6 \\
Gumble Tree-LSTM \cite{gumbel}     & 100D &   82.6   \\
Gumble Tree-LSTM \cite{gumbel}$^{\ast}$     & 100D &   82.3 $\pm$ 0.1   \\
Tree-LSTM + PPO \cite{ppo}        & 100D &    \textbf{84.3 $ \pm$ 0.3}  \\
Ours(Attention)                       & 100D &       83.3 $\pm$ 0.1\\
\midrule
SPINN \cite{Bowman}     & 300D &      83.2 \\
NTI \cite{Munkhdalai} & 300D &      84.6\\
Gumble Tree-LSTM \cite{gumbel}     & 300D &    85.6  \\
Gumble Tree-LSTM \cite{gumbel}$^{\dagger}$    & 300D &   83.7   \\
Gumble Tree-LSTM \cite{gumbel}$^{\mathsection}$    & 300D &     84.9 $\pm$ 0.1 \\
Gumble Tree-LSTM \cite{gumbel}$^{\ast}$    & 300D &     84.4 \\
Tree-LSTM + PPO \cite{ppo}     & 300D &     85.1 $\pm$ 0.2 \\
Ours(Attention)                      & 300D &  \textbf{ 85.9 $\pm$ 0.1}    \\
\bottomrule
\end{tabular}
\end{adjustbox}
\label{tab:snli}
\end{table}

\subsection{Datasets}
\begin{itemize}
    \item \textbf{SNLI} : SNLI \cite{snli} is a natural language inference task, also known as textual entailment, that classifies two sentences with the labels \textit{entailment}, \textit{contradiction} and \textit{neutral}. SNLI is a collection of 570k manually labelled pairs of sentences. We choose training instances where the combined length of premise and hypothesis sentences is less than 120. Our training set consists of 549367 sentences and 9800 each in the validation and test sets.
    \item \textbf{SST} : SST \cite{sst} is a sentiment Treebank that also serves as a classification dataset. It includes fine-grained sentiment labels for 215,154 phrases in the parse trees of 11,855 sentences extracted from movie reviews.  
    This dataset has two labelling schemes: the labelling scheme SST-2 consists of  binary labels (negative, positive) whereas SST-5 consists of  fine-grained sentiment labels (5-way classification).
    \item \textbf{Quora Question Pairs (QQP)} : QQP is a question pair dataset from the popular website Quora\footnote{Quora is a question answering website where users ask questions and other users respond. }
    The aim is to determine whether a pair of questions are duplicates, that is, whether they seek the same answer. QQP dataset consists of 404,290 question pairs. We use $12\%$ of the randomly sampled dataset as the test set. Of the given question pairs, there are 63.08\% with a negative label, and 36.92\%instances with a positive  labels~\cite{DBLP:journals/corr/abs-1907-01041}. The test set has positive to negative label ratio of $\sim$ 0.55.
    \end{itemize}
\subsection{Parameter Settings}
For all the experiments in the paper, we follow the experimental protocol of \cite{gumbel}. We ran SNLI experiments with a batch size of 32; for the 300-dimensional experiments, we use the glove embeddings with a vocabulary size of 840B as an input word embedding.  Whereas for the 100 dimensional experiments, we use glove embeddings over a vocabulary of size 6B.  The dropout rate is set to $0.13$. For the SST experiments, the batch size is set to 64 and the dropout is set to $0.5$ while the word vector dimension set is set at $300$. Models are optimized using stochastic gradient descent with the Adam optimizer~\cite{kingma2014adam} with an initial learning rate of $0.5$. We use accuracy as a metric to evaluate our results.

\section{Results}
\subsection{Natural Language Inference}

We compare our classification accuracy on the SNLI dataset in Table~\ref{tab:snli}. We observe that our approach outperforms the Gumbel Tree LSTM by a margin of $+1$ accuracy score for 100D and $+1.5$ for 300D. Our method performs lower than \cite{ppo} in a 100D setting; however, the attention-based model outperforms by a margin of $+0.7$ in the 300D setting. The standard deviation of the \cite{ppo} is higher than our approach. It seems that gains are marginal on the end-task but we show in Section \ref{sec:quan} that our learned trees are balanced and far superior than previous approaches. We also observe that our approach converges in a time comparable to the Gumbel Tree-LSTM.


\subsection{Sentiment analysis}
We evaluate our model on a sentiment analysis task using the SST dataset. Sequence models outperform the recursive model by a significant margin. These models are pre-trained on the larger dataset and fine-tuned on sentiment analysis tasks. The performance of these models can be attributed to pre-training rather than learning any syntax. However, our approach outperforms Tree-LSTM based approaches substantiating the benefits of attention for sentiment analysis task. \par

\begin{table}[!h]
\caption{Results on SST2 and SST5 datasets. $^\mathsection$: results taken from \cite{williams2018latent}. PPO: Proximal Policy Optimization }
\centering
\begin{adjustbox}{width=0.98\linewidth}
\begin{tabular}{@{}lcc@{}}
\toprule
Model & SST-2 & SST-5\\
\toprule
BYTE mLSTM \cite{Radford}   & \textbf{91.8} & 52.9 \\
CoVe \cite{McCann}   & 90.3 & 54.7 \\
biLM \cite{Peters}     & - &   \textbf{54.7}  \\
\midrule
RNTN \cite{socher-etal-2013-recursive}     & 85.4 &   45.7  \\
parse Tree-LSTM \cite{Tai}        & 88.4 &    51.0  \\
NTI \cite{Munkhdalai}        & 89.3 &    53.1  \\
Tree-LSTM + Dynamic Batching \cite{Looks}        & 89.4 &    52.3  \\
\midrule
RL-SPINN \cite{yogatama2016learning} & 86.5 & -\\
Gumble Tree-LSTM \cite{gumbel} & 90.7 & 53.7\\
Gumble Tree-LSTM \cite{gumbel} $^{\mathsection}$ & 90.3 $\pm$ 0.5 & 51.6 $\pm$ 0.8\\
Tree-LSTM + PPO \cite{ppo} & 90.2 $\pm$ 0.2 & 51.5 $\pm$ 0.4\\
\midrule
Ours(Attention)                       &  \textbf{90.7 $\pm$ 0.2} &      51.7 $\pm$ 0.1  \\
\bottomrule
\end{tabular}
\end{adjustbox}

\label{tab:sst}
\end{table}

\subsection{Semantic Relatedness}
We evaluate our model on duplicate questions task on the QQP dataset. In Table \ref{tab:qqp}, we compare our model against BiLSTM (Bi-directional LSTM), Bi-LSTM with attention and Gumbel Tree-LSTM. Our model performs better than Gumbel tree LSTM by a margin of $+0.5$ and Bi-LSTM based models. Due to improved learning of intermediate structures, our model outperforms previous approaches on the QQP dataset.
\begin{table}[h]

\caption{Results on QQP dataset. $^\ast$: results obtained by keeping same hyperparameters as SNLI on code available for \cite{gumbel} }
\centering
\begin{adjustbox}{width=0.7\linewidth}
\begin{tabular}{@{}lcc@{}}
\toprule
Model  & Macro-F1 \\
\toprule

BiLSTM &83.6\\
BiLSTM + Attention  &84.4\\
Gumble Tree-LSTM \cite{gumbel}$^{\ast}$  &84.9\\
Ours(Attention) &\textbf{85.4} \\
\bottomrule
\end{tabular}
\label{tab:qqp}
\end{adjustbox}
\end{table}

\begin{table*}[!]
\centering
\caption{F1-scores on the MultiNLI development set with respect to strict left- and right- branching trees and with respect to Stanford parser. 
\textit{Macroavg. Depth} is the average height of the tree. All reported numbers are maximum F1-score across each model.}
\begin{adjustbox}{width=0.8\textwidth}
\begin{tabular}{lccccc}
\toprule
Model & \multicolumn{1}{c}{Left Branching} & \multicolumn{1}{c}{Right Branching} & \multicolumn{1}{c}{Stanford Parser} & \multicolumn{1}{c}{Macroavg. Depth} \\
\midrule
300D Gumbel Tree-LSTM& 35.6 & \textbf{40.3} & 25.2 & 4.2\\
~~w/o Leaf GRU &  32.3&
39.9 & 29.0& 4.7\\
300D RL-SPINN & 96.6&  15.8 & 19.0& 8.6\\
~~w/o Leaf GRU & \textbf{99.8}&  11.1& 18.2&8.6\\
Ours Approach & 28.5& 35.0& \textbf{31.3} & 4.7\\
~~w/o Leaf GRU& 32.1& 14.4& 31.0 &\textbf{ 5.3}\\
\bottomrule
\end{tabular}
\end{adjustbox}
\label{tab:quantMultiNLI}
\end{table*}

While our gains on end task for SNLI, SST and QQP are modest, in Section~\ref{sec:qual} we present how our approach is a lot more explainable than existing approaches in two ways: (i) The way the meaning of the parts are composed in our approach vis-a-vis word2vec; our approach has significantly better discriminative ability over word2vec in distinguishing acceptable and non-acceptable answers. (ii) The syntactic composition of parts by our weakly-supervised approach is remarkably better than that obtained using Gumbel tree LSTM. In Section~\ref{sec:quan} we also present a quantitative evaluation of both components, {\em viz.}, (i) meaning of parts, and (ii) the syntactic composition of the parts into meaningful parse trees.

\begin{table*}[h!]
\centering

\caption{Toy examples for finding similar sentences. Scores represented here are cosine similarity scores between the pair of sentences. {\color{green}A}: acceptable answer, {\color{red}N}: not acceptable answer }
\begin{adjustbox}{width=0.65\textwidth}
\begin{tabular}{lcccc}
\toprule
Sentence & Similarity Label & Word2Vec & Gumbel & Ours \\
\midrule
Main : I like to drink orange juice.&&&&\\
\midrule
I love orange juice. & {\color{green}A}  & 0.94 & 0.73 & 0.83\\
I like drinking orange juice. & {\color{green}A}  & 0.90 & 0.82 & \textbf{0.93}\\
I like to eat oranges. & {\color{red}N}  & 0.97 & 0.85 & 0.61\\
I do not like orange juice. & {\color{red}N}  & 0.95 & 0.71 & 0.57\\
\midrule
\midrule
Main : The weather is very hot today.&&&&\\
\midrule

The temperature is very high.&{\color{green}A} &0.85&0.41&0.55\\
It is too hot today. & {\color{green}A}  &0.94&0.70&\textbf{0.77}\\
Today is very cold. & {\color{red}N} &0.94&0.36&0.43\\
The weather is not very high.& {\color{red}N}&0.94&0.37&0.40\\
The weather is not very hot.& {\color{red}N}&0.98&0.50&0.47\\
\bottomrule
\end{tabular}
\end{adjustbox}
\label{tab:simscore}
\end{table*}

\subsection{Quantitative Evaluation over Trees}
\label{sec:quan}

While our models perform better over NLI and semantic relatedness tasks and are comparable over sentiment analysis, we now present how our model offers remarkably better explainability over state-of-the-art approaches. We now quantitatively evaluate the `well-formedness' of our latent tree structures and compare them with tree-structures inferred by the other recent approaches as surveyed in \cite{williams2018latent}. We also present a qualitative analysis of trees induced by our approach in Section~\ref{sec:qual}.

Table \ref{tab:quantMultiNLI} shows parsing performance on the MultiNLI development set for our model. Following the usual practice, we measured automatically generated trees on three different settings: a) when trees are strictly left-branching, b) strictly right-branching, and c) 
trees available in the dataset generated by the Stanford parser. We observe that RL-SPINN\cite{yogatama2016learning} is biased towards left-branching trees, whereas the Gumbel Tree-LSTM method produced better-balanced trees with a slight preference for right-branching trees. Our approach learns better grammar rules equally for left and right branch trees. Further, our model learns far better Stanford style grammar than Gumbel Tree-LSTM and RL-SPINN.
It is to be noted that trees obtained from the Stanford PCFG parser \cite{stanford_parser} are not gold-label trees but generated by Stanford parsing algorithm. Hence, the trees may not be completely accurate with respect to gold-label trees, but it is a regular practice to compare parse trees.
\par
\textit{Macro average depth} is another metric to understand the efficacy of learned trees. It is the average path length from root to any given leaf node(here, a word). For the Stanford Parser trees, the macro average depth is close to $5.7$. RL-SPINN method produces highly skewed and left-branch trees suggesting that it attuned to produce left-branching trees. Our numbers are closer to the baseline than the trees produced by Gumbel Tree-LSTM. The evaluation confirms our hypothesis that the principle of compositionality is helpful while learning the latent structure of trees. 
\par
Additionally, Table \ref{tab:quantMultiNLI} demonstrates that merging decisions during parsing is more informative than Gumbel Tree LSTM due to the proposed attention mechanism.
\subsection{Qualitative Analysis} \label{sec:qual}

\begin{figure*}[!h]
    \centering
    \includegraphics[width=0.8\textwidth]{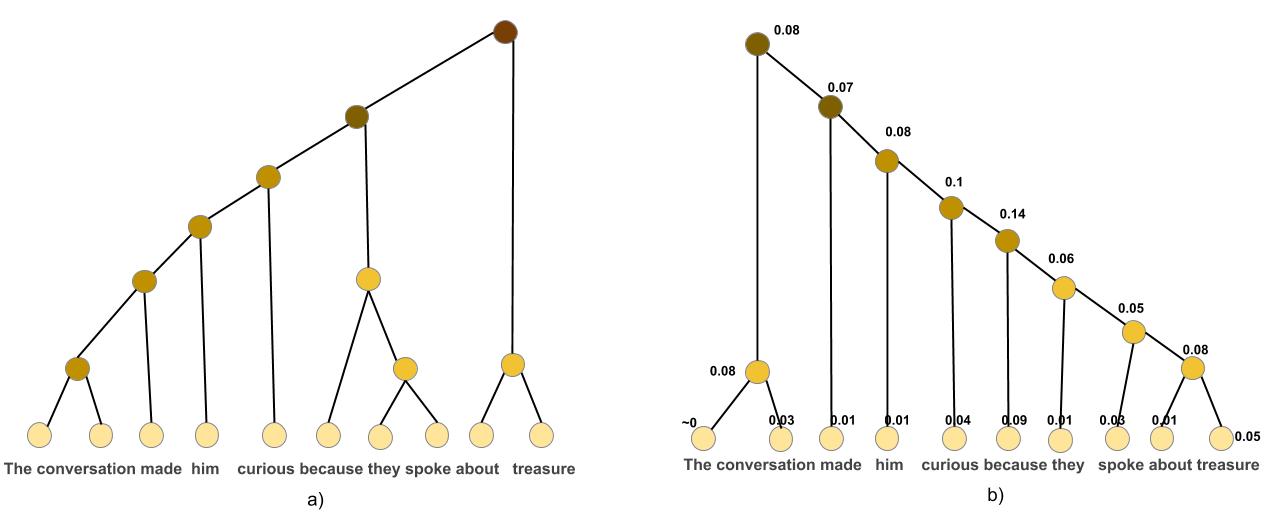}
    \caption{Comparison of learned intermediate parse tree structure of a) Gumbel Tree LSTM with b) our approach on a example from the MultiNLI dataset. The numbers in b) are attention scores for the compositions that serve as the coefficients over which the final sentence representation is generated.} 
    \label{fig:SNLI}
\end{figure*}
\begin{figure*}[!t]
\begin{tabular}{lll}
\hline
{\includegraphics[width=54mm]{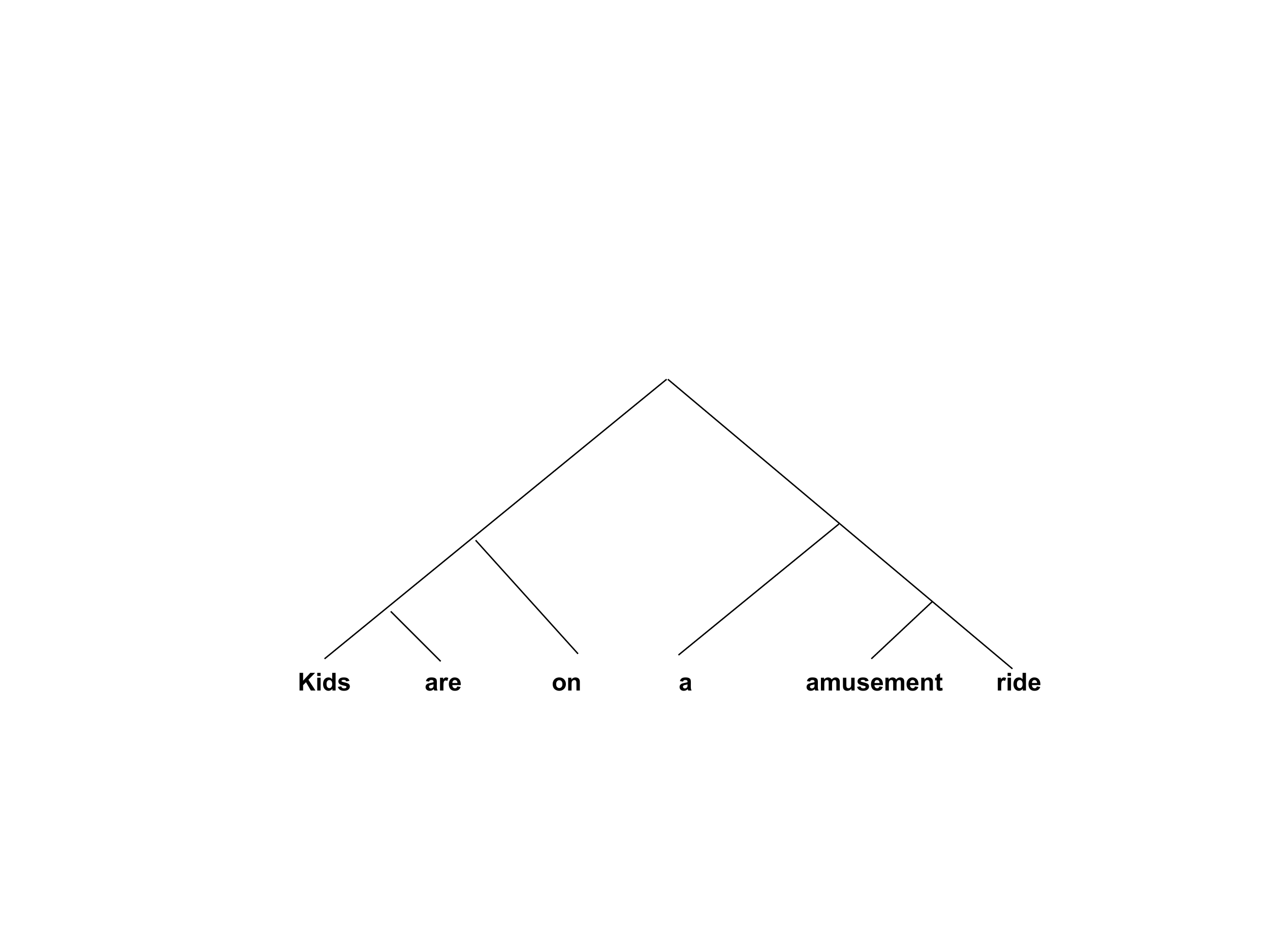}}
     {}
&
{\includegraphics[width=54mm]{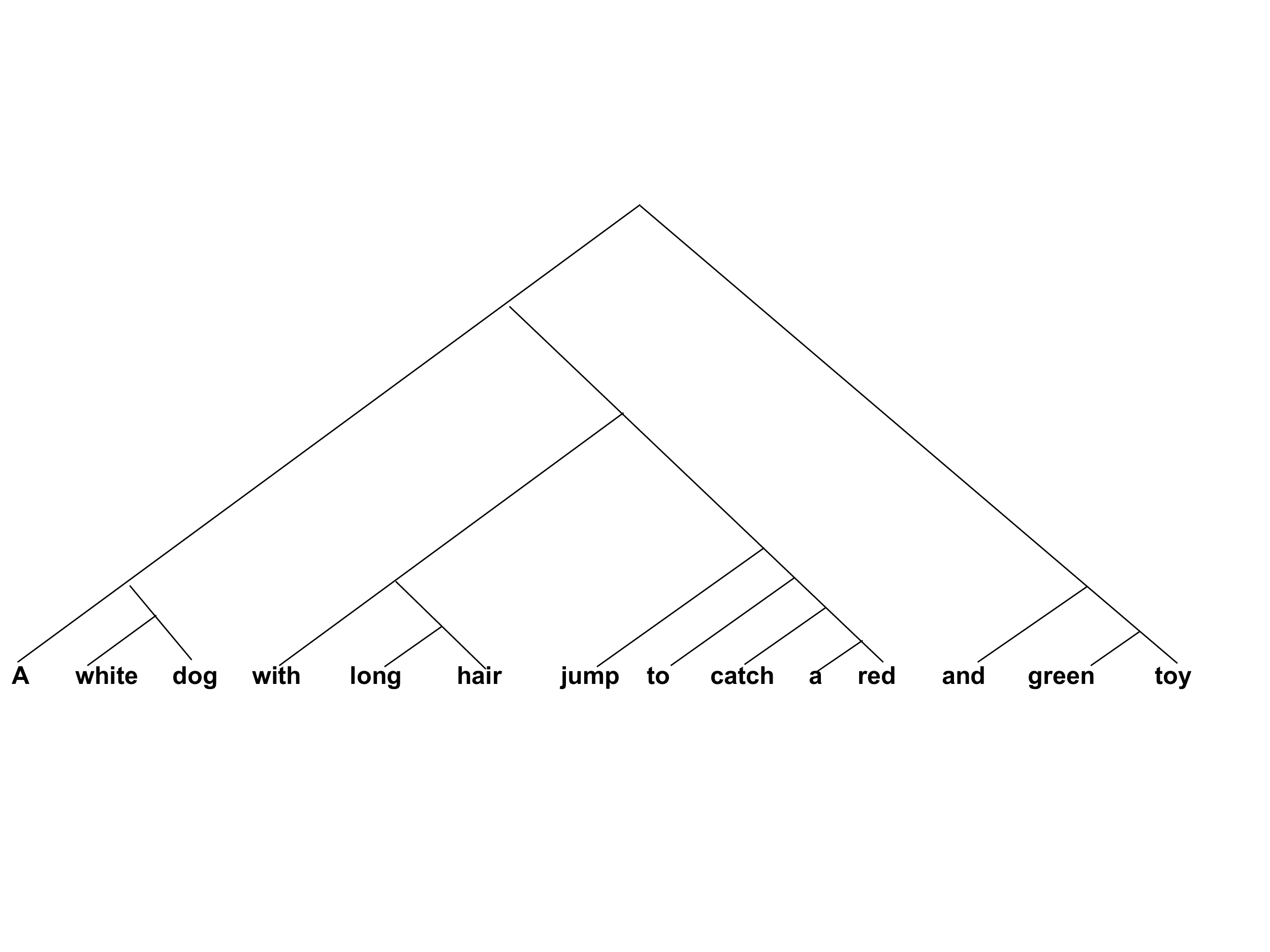}}
     {}
&
{\includegraphics[width=54mm]{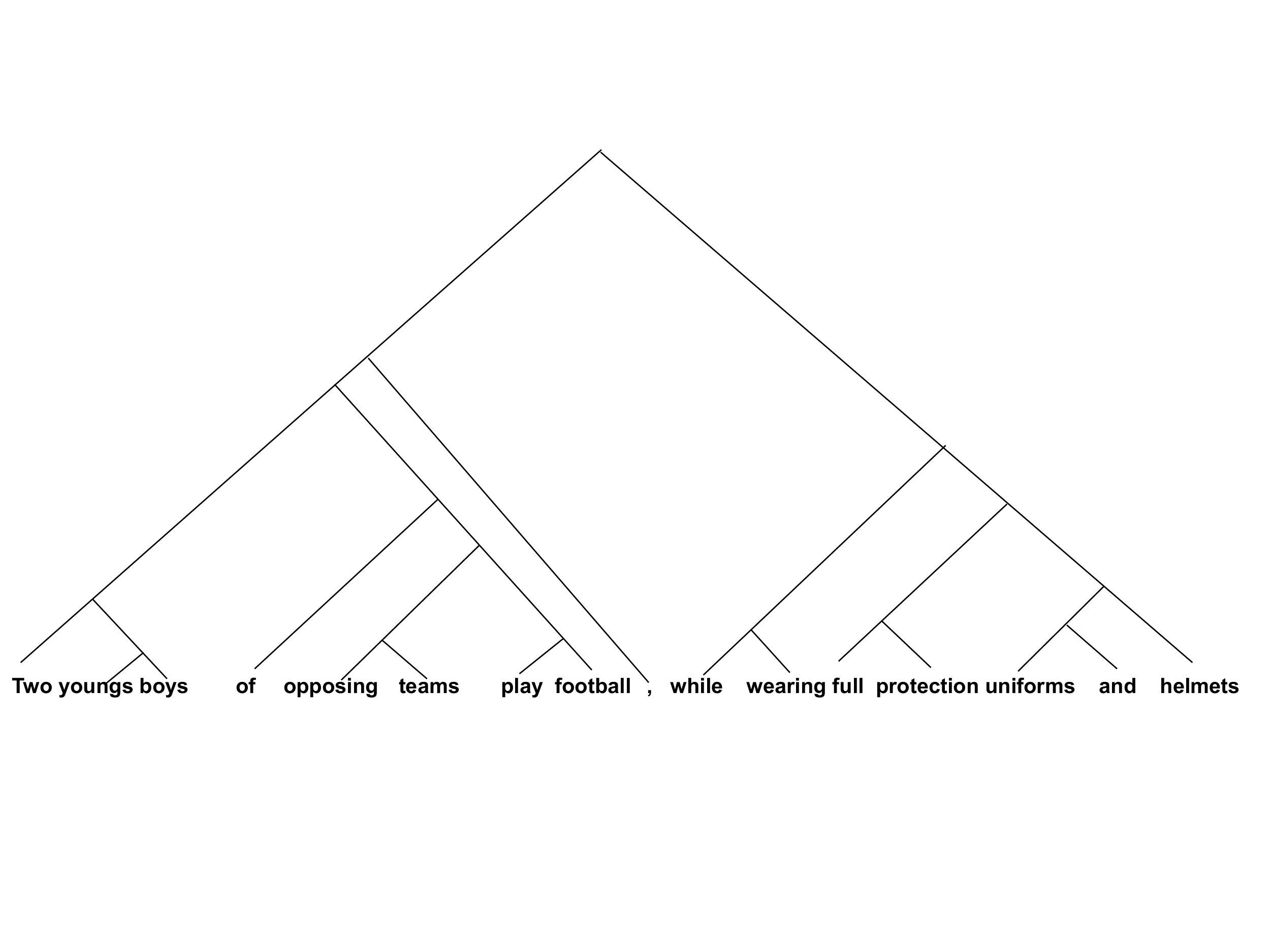}}
     {}
\\

{\includegraphics[width=54mm]{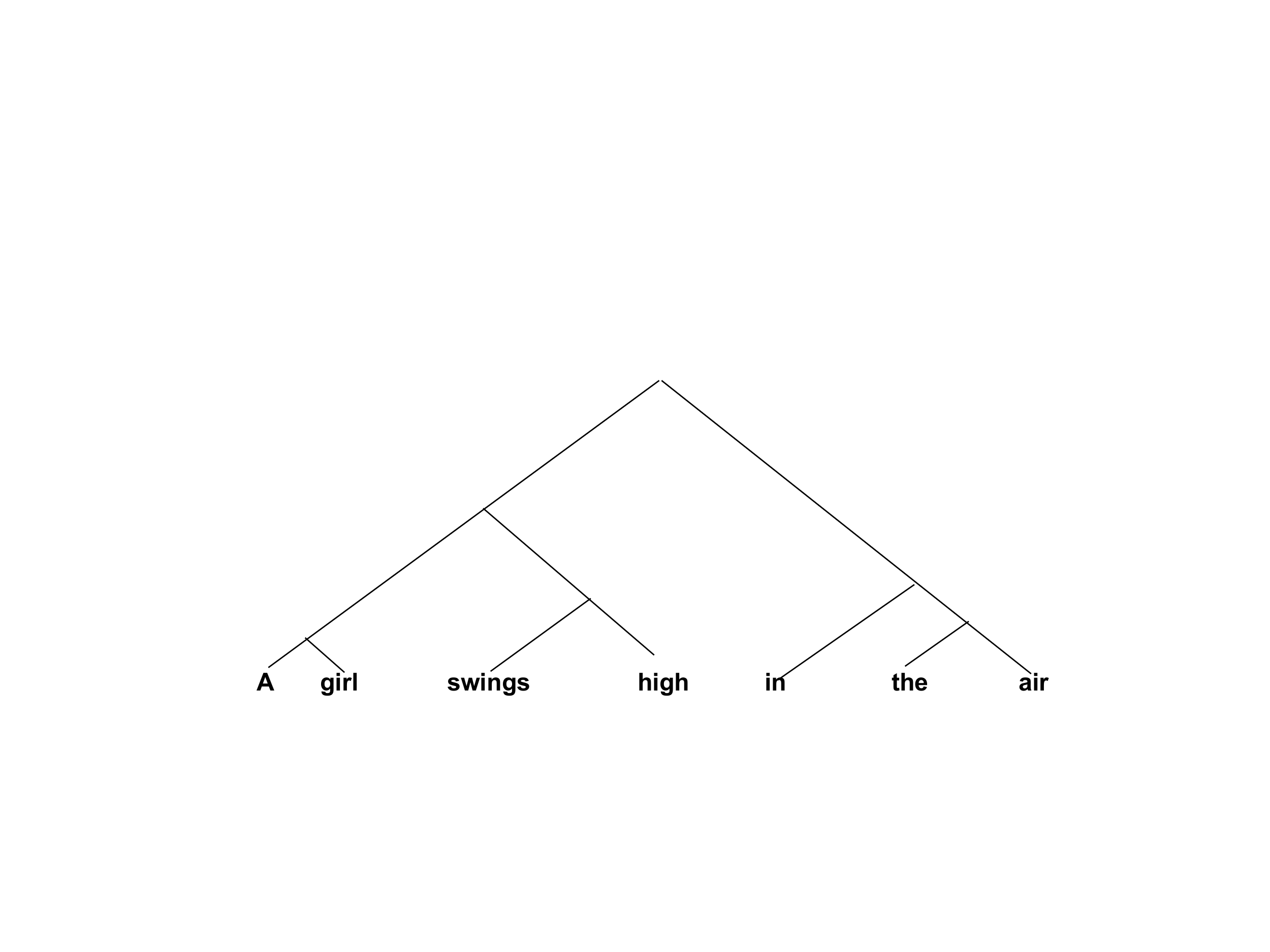}}
     
&
{\includegraphics[width=54mm]{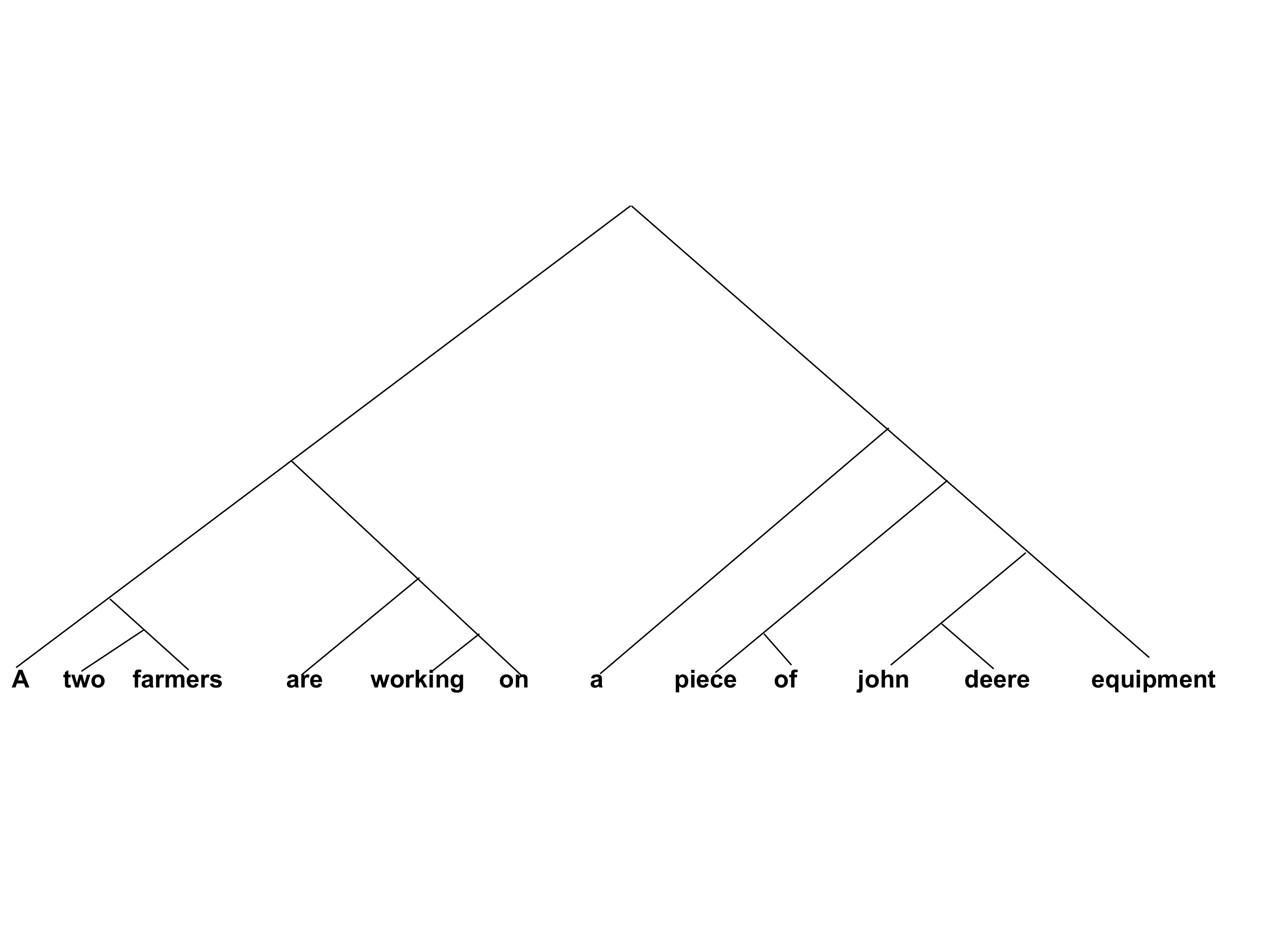}}
     {}
&
{\includegraphics[width=54mm]{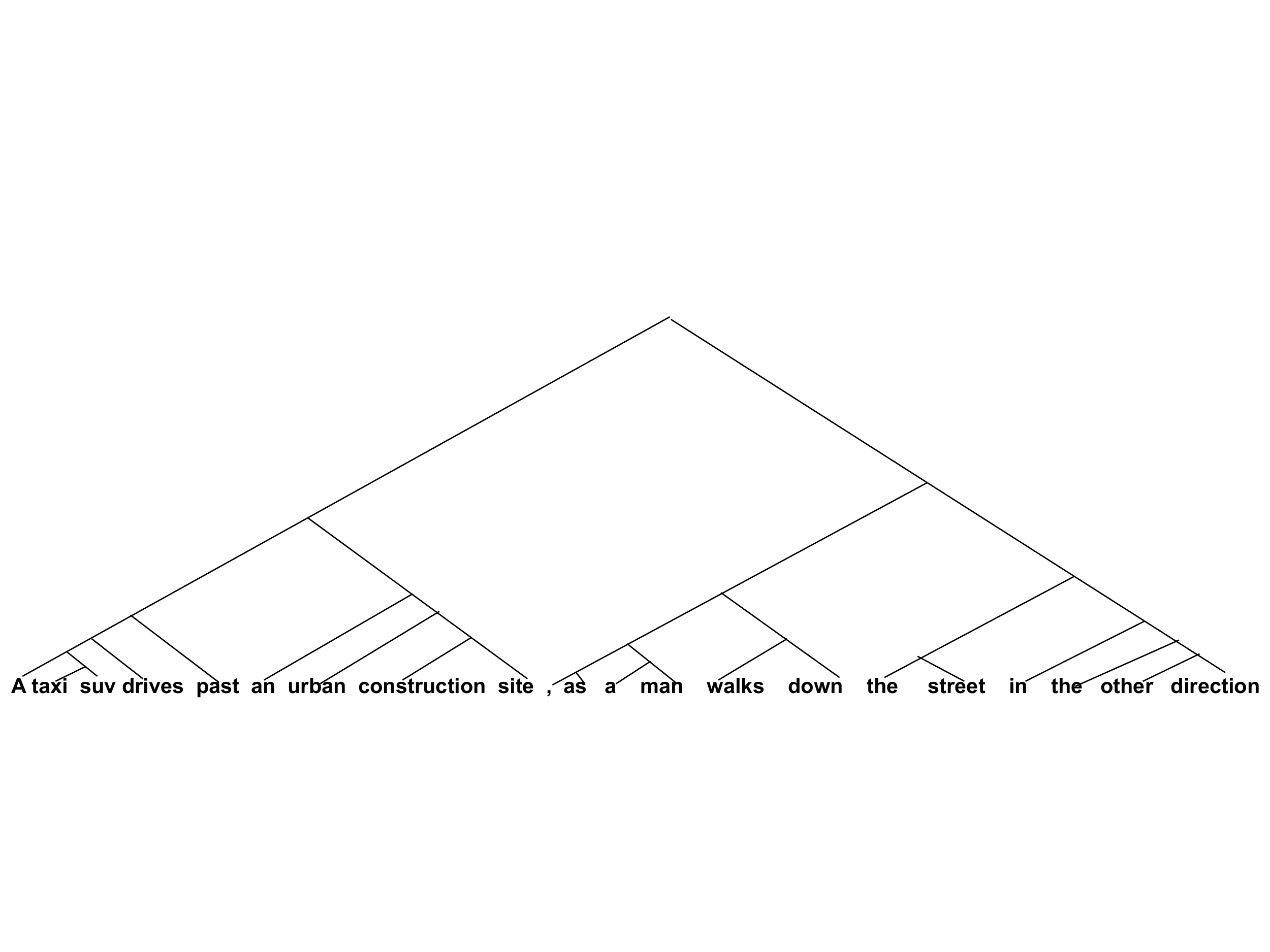}}
     {}
\\
\hline
\end{tabular}
\caption{Shows induced example trees for the small length sentences, Medium Length Sentences and Large length/Complex sentences From L $\rightarrow$ R}
\label{fig:trees}
\end{figure*}

We present induced tree structures learned on different examples from the SNLI dataset in Figure \ref{fig:SNLI} and \ref{fig:trees}. In these examples, Gumbel Tree LSTM produces ambiguous and non-meaningful structures while our model generates syntactically and semantically meaningful parse trees. Interestingly, our model can capture complex sentences producing visually plausible tree structures.
Figure \ref{fig:trees} shows the learned parse trees for three categories of sentences split on the sentence length from SNLI corpus. Small sentences are of length $\le 10$, medium sentences are of  length $\le 15$, and large/complex sentences are of  length $\le 20$ and  include in between punctuation. Evidently, our approach produces grammatically correct parse trees for big sentences as well.

\par

Drawing inspiration from \cite{gumbel}, we analyze the effectiveness of learned embeddings using cosine similarity of related sentences and unrelated sentences. In Table \ref{tab:simscore}, we compare our method with embeddings obtained from the average pooling of word vectors \cite{mikolov2013efficient} and Gumbel tree LSTM. Word2vec embedding gave high similarity to each sentence pair due to the occurrence of similar words. For the sentence \texttt{I like to drink orange juice}, sentence embeddings learned from Gumbel tree LSTM give a high score to both acceptable and non-acceptable answers, whereas our method gives clear separation between acceptable and non-acceptable answers. Similarly, for the second sentence \texttt{The weather is very hot today}, the Gumbel tree LSTM gives a high score to the non-acceptable answer. Instead, our method disambiguates between similar-looking sentences but non-acceptable answers and has a clear separation between acceptable and non-acceptable answers.
\par
\section{Conclusion}
In this paper, we proposed an attention mechanism over Gumbel Tree-LSTM to learn hierarchical structures of natural language sentences. Our model introduces the attention over composition query vector such that constituents are weighted and merged according to its latent part of speech information. We demonstrate the advantage of our method on three tasks: natural language inference, semantic relatedness, and sentiment analysis. We show that our approach outperforms recent state-of-the-art RvNN-based models on all three tasks. Further, we perform an extensive quantitative and qualitative analysis of the induced parse trees and observe that learned induced trees are more explainable, semantically meaningful, and grammatically correct than recent RvNN models. We also observed that the trees learned from the multitask experiments were generalized, and in the future, we would like to explore that direction.

\section{Acknowledgements}
Ayush Maheshwari is supported by a Fellowship from Ekal Foundation (www.ekal.org). We are also grateful to IBM Research, India (specifically the IBM AI Horizon Networks - IIT Bombay initiative) for their support and sponsorship. Manjesh K. Hanawal would like to thank the support from INSPIRE faculty fellowship from DST, Government of India, and Early Career Research (ECR) Award from SERB. 
\bibliographystyle{IEEEtran}
\bibliography{main}
\end{document}